# AtrousMamaba: An Atrous-Window Scanning Visual State Space Model for Remote Sensing Change Detection

*Tao Wang [a,b], Tiecheng Bai [a], Chao Xu [b], Bin Liu [b], Erlei Zhang [b], Jiyun Huang [b], Hongming Zhang [b,\*]*

[a] *College of Information Engineering, Tarim University, Alaer 843300, China*
[b] *College of Information Engineering, Northwest A&F University, Shaanxi 712100, China*

A B S T R A C T

Recently, a novel visual state space (VSS) model, referred to as Mamba, has demonstrated significant progress in modeling long sequences with linear complexity, comparable to Transformer models, thereby enhancing its adaptability for processing visual data. Although most methods aim to enhance the global receptive field by directly modifying Mamba's scanning mechanism, they tend to overlook the critical importance of local information in dense prediction tasks. Additionally, whether Mamba can effectively extract local features as convolutional neural networks (CNNs) do remains an open question that merits further investigation. In this paper, We propose a novel model, AtrousMamba, which effectively balances the extraction of fine-grained local details with the integration of global contextual information. Specifically, our method incorporates an atrous-window selective scan mechanism, enabling a gradual expansion of the scanning range with adjustable rates. This design shortens the distance between adjacent tokens, enabling the model to effectively capture fine-grained local features and global context. By leveraging the atrous window scan visual state space (AWVSS) module, we design dedicated end-to-end Mamba-based frameworks for binary change detection (BCD) and semantic change detection (SCD), referred to as AWMambaBCD and AWMambaSCD, respectively. Experimental results on six benchmark datasets show that the proposed framework outperforms existing CNN-based, Transformer-based, and Mamba-based methods. These findings clearly demonstrate that Mamba not only captures long-range dependencies in visual data but also effectively preserves fine-grained local details.

## 1. Introduction

Remote sensing change detection (RSCD) refers to the process of detecting surface changes on the Earth across different time periods by utilizing satellite or aerial imagery. This technique plays a vital role in various applications such as environmental monitoring, urban expansion analysis, and agricultural land use assessment (Peng et al., 2025). With the rapid progress in remote sensing technologies, a vast amount of imagery with varying spatial resolutions and sensor modalities has become available, thereby intensifying the need for automated methods to identify land cover changes from bi-temporal image pairs (Zhu et al., 2024b).

According to the semantic label requirements in the output change maps, change detection tasks are generally divided into binary change detection (BCD) and semantic change detection (SCD) (Ding et al., 2022; Zhang et al., 2020a). The objective of BCD is to differentiate changed pixels from unchanged ones within bi-temporal imagery using binary classification labels (Chen et al., 2021; Fang et al., 2021). In comparison, SCD poses a greater challenge, as it not only detects change regions but also assigns semantic categories to the changes, thereby offering fine-grained "from-to" transitions at the pixel level (Daudt et al., 2019; Peng et al., 2021; Zheng et al., 2022). Most existing approaches are tailored exclusively for either BCD or SCD, and it remains an open question whether a unified framework can effectively improve performance across both tasks.

\* *Corresponding author.*
E-mail address: zhm@nwsuaf.edu.cn

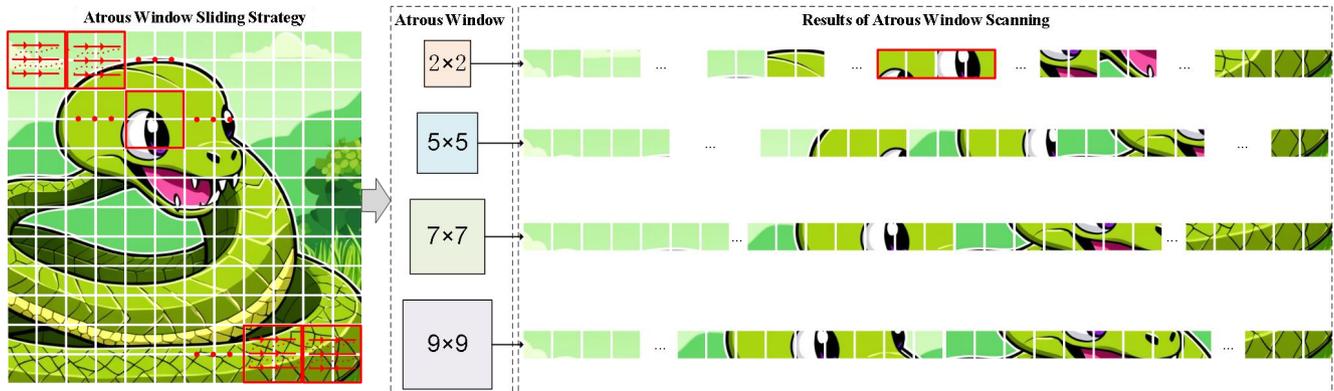

**Fig. 1.** atrous window scanning applies four different rates to partition the image into four groups of windows with varying sizes. Within each group, the image patches from all windows are unfolded and concatenated into sequences. These four groups of patch sequences are subsequently processed in parallel using independent S6 modules.

To extract intricate patterns from high-resolution optical remote sensing images, two mainstream architectures—Convolutional Neural Networks (CNNs) and Vision Transformers (ViTs)—are widely adopted for both BCD and SCD tasks, (Cui and Jiang, 2023; Daudt et al., 2018; Daudt et al., 2019; Ding et al., 2024; Dong et al., 2024; Li et al., 2022). A notable drawback of CNNs lies in their limited ability to model global contextual information, which stems from their inherently constrained receptive fields. In contrast, ViTs are generally more effective at capturing global dependencies, owing to the incorporation of self-attention mechanisms (Dosovitskiy et al., 2020; Vaswani et al., 2017), that allow for dynamic weighting and unrestricted receptive field coverage. Nevertheless, this advantage comes at a cost: the self-attention operation scales quadratically with input size, resulting in considerable computational burden when processing high-resolution data. To alleviate this issue, researchers have proposed more efficient attention variants by limiting window sizes or adjusting computation strides (Dong et al., Jun. 2022; Liu et al., Oct. 2021; Zhang et al., Feb. 2022). Despite their improvements, these approaches inherently compromise between computational efficiency and the ability to model long-range spatial interactions.

Structured state space sequence models have recently gained attention as a compelling alternative for sequence modeling tasks. The structured state space sequence model (S4) (Gu et al., 2021) introduces an innovative parameterization that unifies continuous-time, recurrent, and convolutional formulations of state space systems, allowing for principled and efficient modeling of long-range dependencies. To overcome certain limitations of S4, Mamba (Gu and Dao, 2023) incorporates dynamic weight modulation during sequence propagation, substantially expanding the effective receptive field. These models, characterized by their linear computational complexity, have demonstrated strong performance in handling long sequences, achieving notable success in areas such as computer vision (Liu et al., 2024; Xiao et al., 2024; Zhu et al., 2024a), and medical image segmentation (Ma et al., 2024a; Ruan and Xiang, 2024). Nonetheless, the scanning mechanisms used in these architectures often lead to spatial discontinuities between adjacent pixels, thereby impeding smooth information propagation. Moreover, when applied to remote sensing imagery, these models still exhibit shortcomings, highlighting the need for further exploration of their effectiveness across different change detection subtasks.

This study introduces AtrousMamba, an efficient architecture built upon the Mamba framework, specifically tailored for RSCD tasks. The proposed approach harnesses Mamba's capabilities to effectively model both local and global contextual dependencies, while also capturing spatio-temporal patterns from multi-temporal inputs. Through the integration of multi-scale contextual features, the model facilitates accurate identification of diverse land-cover transitions and demonstrates strong performance across both BCD and SCD subtasks. Specifically, as illustrated in Fig.1, the proposed module utilizes an atrous window scanning visual state space model with multiple dilation rates, partitioning the image into four groups of windows at different scales. Within each group, the image patches from all windows are unfolded and arranged into sequences along a horizontal traversal path. These four groups of patch sequences are processed in parallel using independent S6 modules, ensuring that labels within the same 2D semantic region are processed together, while establishing connections between semantic labels across different windows. By progressively expanding the receptive field, the model is able to capture a greater amount of local feature information.

In summary, the key contributions of this study are as follows:

1) We present a novel scanning methodology for state space models (SSMs), called the atrous window scanning visual state space (AWVSS), which integrates four selective scan mechanisms with varying window rates, enhancing the model's ability to capture detailed local information while preserving global context.

2) Building upon VSS and AWVSS, we propose a customized end-to-end Mamba network tailored for two change detection tasks, namely BCD and SCD, thereby significantly enhancing both model accuracy and computational efficiency.

3) The atrous window scan module (AWSM) is designed to preserve the spatial proximity between adjacent tokens across different window scan sequences, enabling the simultaneous establishment of both global and local receptive fields, while enhancing inter-channel information interaction.

4) We perform extensive experiments on six widely used remote sensing image datasets for BCD and SCD. The results demonstrate that AtrousMamba achieves competitive accuracy compared to state-of-the-art CNN- and Transformer-based methods, and surpasses recently proposed Mamba-based change detection methods.

## 2. Related work

### 2.1. Binary change detection

The siamese network-based change detection method was first proposed in (Zhan et al., 2017). A large number of current change detection approaches adopt the siamese network framework, which enables the parallel encoding of bi-temporal remote sensing imagery. CNNs have been widely utilized in BCD tasks due to their strong ability to extract meaningful features from spatial data, particularly in the context of remote sensing applications. Extensive research efforts have been devoted to enhancing change detection through various techniques, including multilevel feature fusion (Zhang et al., 2020a; Zhao et al., 2023), difference-based representation modeling (Shu et al., 2022; Song et al., 2022), and the incorporation of attention mechanisms (Fang et al., 2021; Wang et al., 2021; Yin et al., 2023).

Three fully convolutional neural network (FCNN) architectures (Daudt et al., 2018) have been proposed for change detection on multi-temporal pairs of Earth observation images, marking the first use of two fully convolutional siamese architectures with skip connections. FDCNN (Zhang et al., 2020b) extracts deep features from RS images using a CNN and builds a two-channel network with shared weights through transfer learning to generate a multiscale and multidepth feature difference map for BCD. SAGNet (Yin et al., 2023) proposes an attention-guided siamese network for BCD that establishes internal links between spatial context, high-level, and low-level features, while simultaneously fusing global and difference information. GAS-Net (Zhang et al., 2023) leverages the self-attention mechanism to enhance contextual learning and feature extraction, and utilizes the foreground-awareness module to strengthen foreground information, thereby facilitating the exploration of relationships between the scene and foreground. Considering that the degradation of local feature details in high-resolution remote sensing (HRRS) images can lead to the incorrect identification of "non-semantic changes" and the incomplete or irregular extraction of boundaries. CF-GCN (Wang et al., 2024c) constructs a spatial interaction graph convolution based on the concept of non-local blocks and incorporates a feature interaction branch to facilitate channel interactions. This enables the capture of the spatial context of each region and its relationship with neighboring regions, allowing for effective inference of changed areas.

The aforementioned approaches have contributed notably to performance gains in change detection by effectively extracting local features. Nevertheless, the intrinsic locality of convolutional architectures limits their ability to model long-range dependencies, which becomes particularly problematic when the changed areas are sparse and scattered within a large background. To address these challenges, an increasing number of studies have investigated the use of attention mechanisms to strengthen feature representations and improve overall detection performance. BIT (Chen et al., 2021) introduces transformers to efficiently model the context within bi-temporal images, representing the input images as a set of tokens and modeling the context in a compact token-based space-time representation. ChangeFormer (Bandara and Patel, 2022) introduces a transformer-based siamese network architecture for BCD using a pair of co-registered remote sensing images. SwinSUNet (Zhang et al., 2022) designs a pure transformer network with a siamese U-shaped structure to extract multiscale features from bi-temporal images, enabling better extraction of global spatial-temporal information. ICIF-Net (Feng et al., 2022) proposes an intra-scale cross-interaction and inter-scale feature fusion network that effectively integrates CNN and Transformer to jointly leverage both local and global features. DMINet (Feng et al., 2023) unifies self-attention and cross-attention in a single module to guide the global feature distribution of each input, fostering information coupling between intra-level representations while simultaneously suppressing task-irrelevant interference.

While previous methods have achieved moderate improvements in change detection, they often face challenges in balancing accuracy with computational complexity. This is largely due to the localized receptive fields of convolutional networks and the high resource demands of Transformer-based models. Most existing approaches prioritize detection accuracy but tend to overlook computational efficiency. Furthermore, they exhibit limited capability in identifying "non-semantic change" regions, which undermines their overall effectiveness. In contrast, the proposed architectures for both BCD and SCD, developed based on Atrous Window Scan Visual State Space (AWVSS) modeling, strike a favorable balance between global context awareness and local detail extraction, enabling highly efficient and effective change detection in remote sensing imagery.

### 2.2. Semantic change detection

Most existing studies in change detection primarily target BCD, which determines the locations of changes within multi-temporal imagery. In contrast, SCD not only identifies changed regions but also assigns land-cover categories to the detected changes, thereby revealing the nature of the transformations. To tackle the SCD task, multitask learning frameworks have been proposed. One typical design employs an encoder with three separate branches, each dedicated to change detection and two semantic segmentation (SS) subtasks, respectively (Daudt et al., 2019). Alternatively, a two-branch encoder structure has been introduced, where shared feature representations are utilized across the CD and SS subtasks (Ding et al., 2022).

SCDNet (Peng et al., 2021) adopts a siamese UNet architecture with shared weights, facilitating effective multi-level feature representation and fusion for bi-temporal images. To capture changes at multiple scales, the encoders incorporate multi-scale atrous convolution units. Furthermore, in the decoding stage, an attention mechanism and deep supervision are introduced to improve feature fusion and mitigate the issue of gradient vanishing. Building upon the novel CNN architecture (SSCD-l), Bi-SRNet (Ding et al., 2022) integrates two siamese semantic reasoning blocks to model the semantic information in each temporal branch. Additionally, a cross-temporal semantic reasoning block is employed to capture temporal correlations, while a semantic consistency loss function is used to align semantic and change representations. ChangeMask (Zheng et al., 2022) decouples the SCD task into a temporal-wise semantic segmentation task and a BCD task, then integrates these two tasks into a general encoder-transformer-decoder framework that exploits semantic-change causal relationship and temporal symmetry.

CNNs exhibit limitations in modeling interactions across different feature layers from a holistic viewpoint, often leading to inaccuracies in detecting changes, particularly within complex scenes containing diverse objects. Recent research efforts have therefore concentrated on enhancing change difference extraction and semantic feature representation by employing techniques such as attention mechanisms and multi-scale feature fusion. MTSCD-Net (Cui and Jiang, 2023) is a multi-task learning approach designed to fully exploit the correlation between the semantic segmentation task and the BCD task by combining CNN and Transformer architectures. SCanNet (Ding et al., 2024) employs a triple 'encoder-decoder' CNN architecture to extract semantic and change features and introduces SCanFormer to explicitly model the 'from-to' semantic transitions between bi-temporal remote sensing images (RSIs). Additionally, it incorporates spatio-temporal constraints aligned with the SCD task to effectively guide the learning of semantic changes. The CdSC (Wang et al., 2024b) network develops a 3-D cross-difference module to explore deep differences within spatiotemporal instance features. Furthermore, an SCE module is introduced to enhance the consistency between the difference features and bitemporal representations. HGINet (Long et al., 2024) leverages graph learning to model the interactions between different feature layers, enhancing detection performance in complex SCD scenarios. Additionally, it employs a cross-learning strategy to improve the identification of unchanged regions.

The majority of current change detection techniques are designed to address either BCD or SCD independently. Conversely, the proposed method exhibits robust generalization capabilities and delivers consistent performance across both BCD and SCD tasks.

*2.3. State space model*

Mainstream foundational models predominantly utilize CNN and Transformer architectures, which are prevalent across both vision and language domains. Nevertheless, the constrained receptive fields inherent to CNNs and the substantial computational demands of Transformers present obstacles to achieving an optimal trade-off between performance and efficiency. State space models (SSMs) (Gu et al., 2021; Smith et al., 2022) address this challenge by modeling sequences in a recurrent manner. The enhanced mechanism, known as Mamba (Gu and Dao, 2023), incorporates weight modulation during propagation, effectively expanding the receptive field and achieving superior results in NLP tasks. Building on this, numerous works have attempted to adapt Mamba for computer vision by applying various predefined methods to convert 2D image features into 1D sequences.

To address the constraints of unidirectional modeling and lack of positional encoding, a novel universal vision backbone, named Vim (Zhu et al., 2024a), has been introduced. Vim incorporates position embeddings to encode spatial information within image sequences and employs bidirectional SSMs to compress visual representations effectively. A single scanning operation faces difficulty in simultaneously capturing dependency information across multiple directions. To overcome this limitation, the Cross-Scan Module (CSM) was incorporated into VMamba (Liu et al., 2024), enabling one-dimensional selective scanning within the two-dimensional image space and thereby facilitating the formation of global receptive fields. However, prior approaches like Vim and VMamba face challenges in effectively capturing dependencies among neighboring pixels within the same semantic region due to the extended spatial distances. In contrast, LocalMamba (Huang et al., 2024) partitions tokens into distinct windows, allowing traversal within each window and thereby improving the ability to capture local dependencies. The GrootVL (Xiao et al., 2024) network dynamically generates a tree topology based on spatial relationships and input features, effectively eliminating the constraints of the original sequence and thereby enhancing the network's representation capabilities.

In remote sensing, RSMamba (Chen et al., 2024b) introduces a dynamic multi-path activation mechanism that overcomes the limitations of unidirectional modeling and positional insensitivity in the vanilla Mamba, achieving superior performance across various remote sensing image classification datasets. RS3Mamba (Ma et al., 2024b) integrates VSS with self-attention mechanisms through a collaborative completion module (CCM). While this integration contributes to improved segmentation accuracy, it also significantly increases the model complexity due to the incorporation of self-attention. The RSM (Zhao et al., 2024) integrates an omnidirectional selective scan module to globally model contextual information in multiple directions, extracting large spatial features across various directions and efficiently performing dense prediction tasks. ChangeMamba (Chen et al., 2024a) adopts the cutting-edge visual Mamba architecture as its encoder, enabling the comprehensive learning of global spatial contextual information from the input images. CDMamba (Zhang et al., 2024) is designed to leverage Mamba's global feature extraction capabilities while enhancing local detail through convolution. Despite achieving promising performance, the method requires an extended training duration of 300 epochs. Prior research has investigated multiple scanning strategies to serialize images and improve Mamba's capacity for visual data interpretation. Nevertheless, whether employing unidirectional or combined scanning directions, these methods generally process entire rows or columns, resulting in excessively long spatial distances when modeling dependencies among neighboring pixels within the same semantic region. Additionally, such strategies induce spatial discontinuities between adjacent pixels, thereby hindering the smooth propagation of information along the sequence.

In contrast, this work investigates Mamba's capability for local feature extraction by employing a progressive atrous window scanning strategy and proposes the AtrousMamba architecture. This framework successfully combines global contextual representation with the preservation of fine-grained local details, leading to substantial improvements in change detection performance.

## 3. Methodology

*3.1. Preliminaries*

*3.1.1. State space models*

SSMs (Gu et al., 2021) process an input sequence $u(t) \in R^L$ through a hidden state representation $h(t) \in R^N$ to produce the output response $y(t) \in R^L$. These models are mathematically formulated using linear ordinary differential equations as follows:

$$\begin{cases} h(t) = Ah(t) + Bu(t), \\ y(t) = Ch(t), \end{cases} \quad (1)$$

where $A \in R^{N \times N}$, $B \in R^{N \times 1}$, and $C \in R^{N \times 1}$ are the weighting parameters, respectively.

For practical implementation, the continuous system is discretized using the zero-order hold method, which transforms the continuous-time parameters $(A, B)$ into their discrete counterparts $(\overline{A}, \overline{B})$ over a sampling interval $\Delta$:

$$\begin{cases} \overline{A} = \exp(\Delta A) \\ \overline{B} = (\Delta A)^{-1}(\exp(\Delta A) - I) \cdot \Delta B \end{cases} \quad (2)$$

This leads Equation (1) to a discretized model formulation as:

$$\begin{cases} h_t = \overline{A}h_{t-1} + \overline{B}u_t, \\ y_t = Ch_t, \end{cases} \quad (3)$$

To improve computational efficiency, the iterative process outlined in Equation (3) can be accelerated by leveraging parallel computation through global convolution, as detailed below:

$$\begin{gathered} y = x \otimes \overline{K} \\ \text{with } \overline{K} = (C\overline{B}, C\overline{AB}, \dots, C\overline{A}^{L-1}\overline{B}) \end{gathered} \quad (4)$$

where $\otimes$ denotes the convolution operation, and $\overline{K} \in R^L$ acts as the kernel of the SSM. This method utilizes convolution to produce outputs over the entire sequence concurrently, thereby improving computational efficiency and scalability.

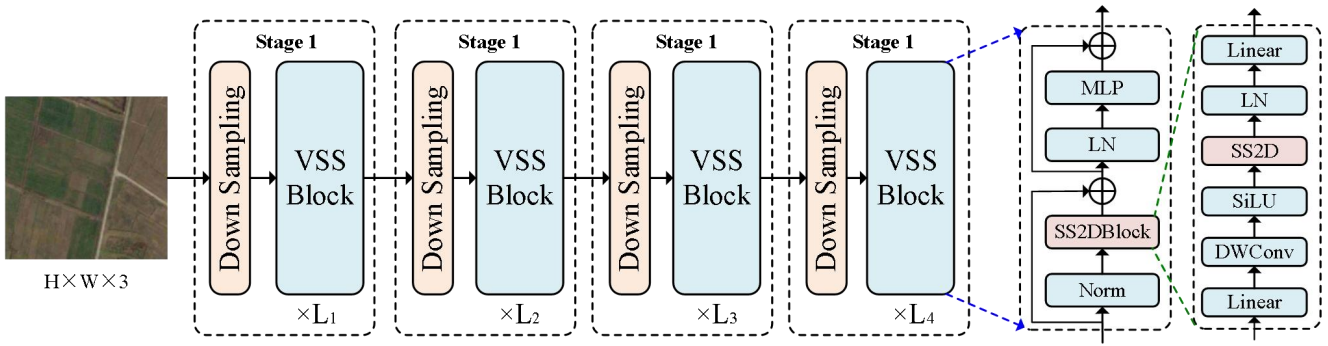

**Fig. 3.** The backbone is based on VMamba, which is employed in our change detection framework.

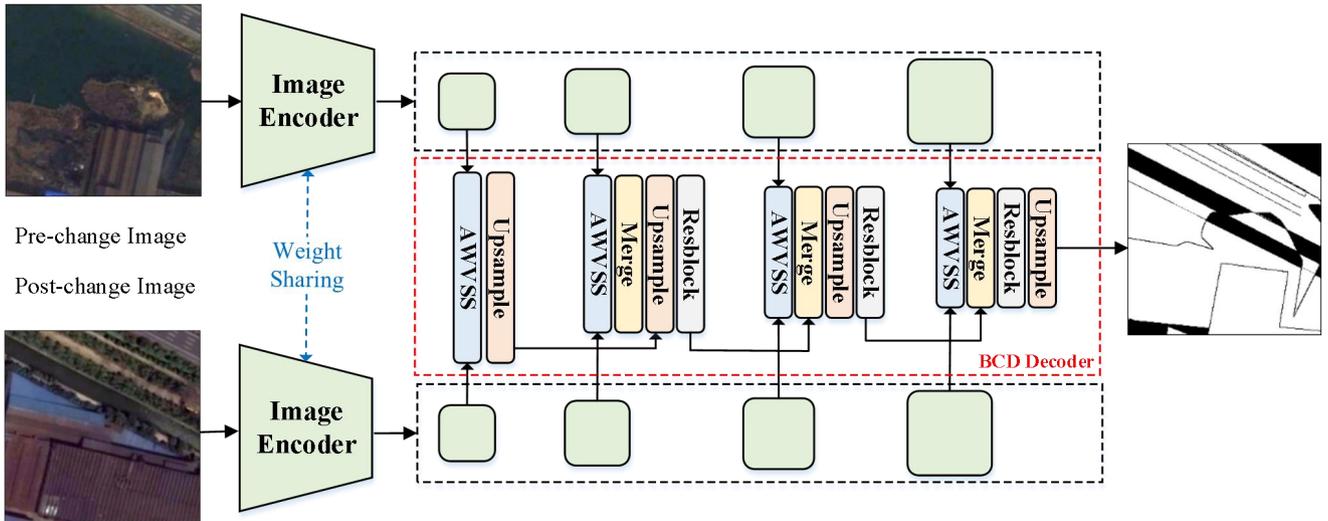

**Fig. 4.** The binary change detection network is built upon the proposed VSS and AWVSS architecture.

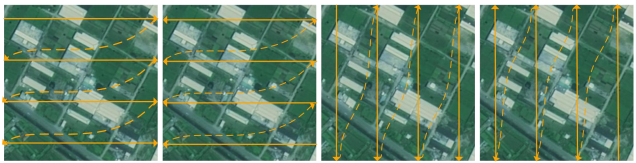

**Fig. 2.** the four scanning paths in CSM

### 3.1.2. Selective state space models

Traditional SSMs, such as S4 (Gu and Dao, 2023), are designed to capture sequential context with linear time complexity. However, their reliance on static parameterization limits their capacity for content-dependent reasoning. To overcome this constraint, selective SSMs—referred to as Mamba (Liu et al., 2024) —introduce dynamic modulation of system matrices A, B, and C, making them responsive to input signals and thus shifting toward a data-driven paradigm. By computing B, C, and the discretization step and Δ directly from the input sequence u(t), Mamba enhances flexibility and sequence-awareness. In addition to preserving linear scalability with respect to sequence length, Mamba also opens new avenues for applications in vision-related tasks.

### 3.1.3. Visual state space model

Building upon Mamba's capability to efficiently model long sequences, VMamba (Liu et al., 2024) proposes a versatile vision backbone that retains the representational power of Vision Transformers (ViTs) while significantly reducing computational overhead. As shown in Fig. 2, VMamba further improves its compatibility with visual data by integrating the Cross-Scan Module (CSM), which enables one-dimensional selective scanning across two-dimensional spatial domains, thereby establishing a global receptive field and enhancing overall performance.

VMamba first partitions the input image into patches using a stem module, producing a two-dimensional feature map with spatial dimensions of H/4 × W/4. The subsequent architecture comprises multiple hierarchical stages, each built with VSS blocks that incorporate the S6 module to facilitate selective scanning tailored for 2D visual inputs. These VSS blocks constitute the core components of VMamba and are preceded by downsampling layers in all stages except the first. Through this design, the model progressively generates multiscale feature representations with spatial resolutions of H/8 × W/8, H/16 × W/16, and H/32 × W/32. The down-sampling operation is realized via patch merging, whereas the detailed architecture of the VSS block is depicted in Fig.3. VMamba is implemented in three configurations: VMamba-Tiny, VMamba-Small, and VMamba-Base. Building upon the VMamba backbone, the encoder structures of the two proposed architectures are depicted in Fig. 4 and Fig. 5. The multiscale features extracted from the four hierarchical stages are subsequently fed into the BCD decoder and the SCD decoder, respectively

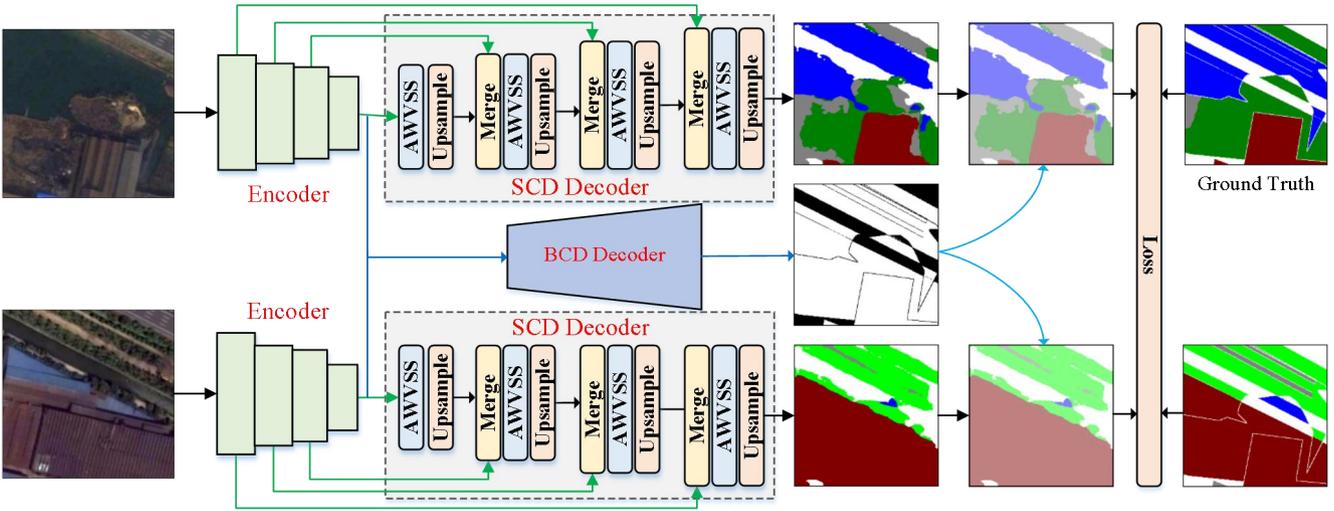

**Fig. 5.** The semantic change detection network is built upon the proposed AtrousMamba architecture.

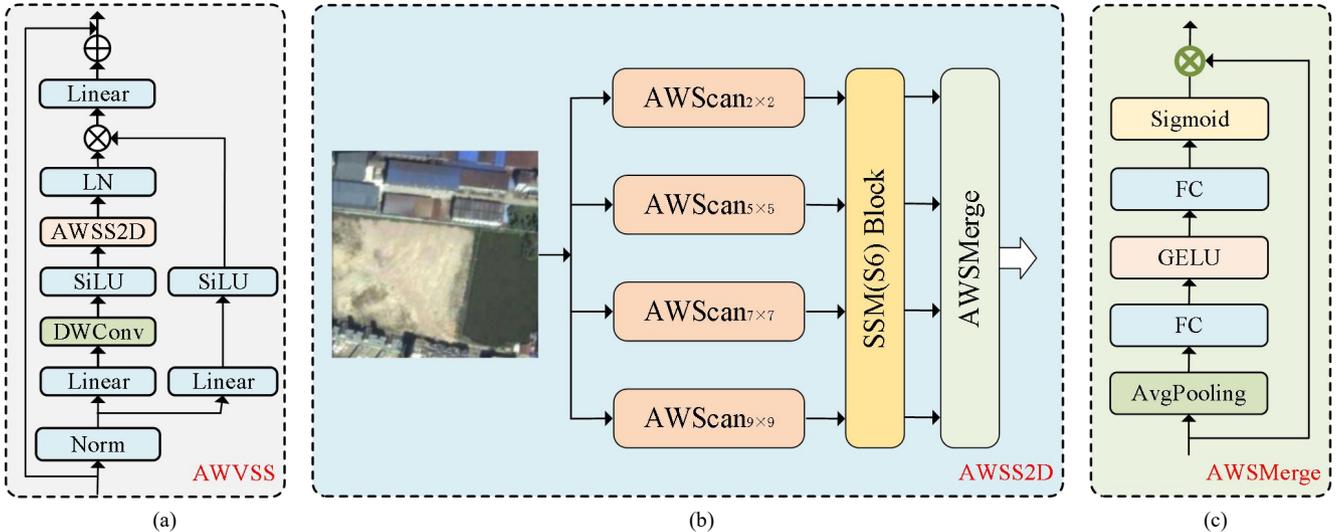

**Fig. 6.** Illustration of the atrous window scanning mechanism. (a) the atrous window visual state space, (b) the atrous window SS2D, (c) the atrous window scan merge module.

### 3.2. Network architecture

We propose an end-to-end Mamba-based network architecture for change detection, leveraging AtrousMamba and VMamba, and name them AWMambaBCD and AWMambaSCD for binary and semantic change detection, respectively. As shown in Fig. 4 and Fig. 5, The encoders for the BCD and SCD networks adopt a weight-sharing siamese architecture based on the VMamba, while the decoders are designed using the AtrousMamba framework. VMamba is a general-purpose visual backbone network built upon the SSMs, characterized by a global receptive field and dynamic weights, which enhance the efficiency of visual representation learning. The AtrousMamba architecture enables the model to capture global contextual information effectively while preserving fine-grained local features, thus enhancing its capability for change detection.

#### 3.2.1. AWMambaBCD:

AWMambaBCD is a dedicated architecture tailored for the BCD task. It begins with a siamese encoder that extracts multi-level features from bi-temporal input images. These features are subsequently fed into a task-specific change decoder, which is constructed based on the AtrousMamba framework. By progressively expanding the receptive field, the decoder effectively captures spatiotemporal dependencies across different feature levels, thereby enabling precise binary change detection. The detailed architecture of the BCD decoder is provided in Section D.

#### 3.2.2. AWMambaSCD

The multitask SCD network employing a triple-branch decoding architecture has been widely adopted in recent studies (Ding et al., 2022). Building upon the AWMambaBCD framework, our proposed AWMambaSCD incorporates three decoder branches: a central branch serving as the change detection (CD) head, and two auxiliary branches (top and bottom) acting as semantic segmentation (SS) heads. The detailed design of the semantic change detection decoder is described in Section D.

### 3.3. Atrous window scanning for visual representations

The current Mamba-based methods (Chen et al., 2024a; Chen et al., 2024b; Ma et al., 2024b; Wang et al., 2024a) inherit a significant limitation from VMamba's image flattening approach, which leads to a loss of locality in change regions and fails to fully utilize spatiotemporal contextual information. To address these issues, we introduce a novel approach, termed AWVSS, as illustrated in Fig. 6 (a), which leverages atrous windows for progressive scanning. The residual network consists of two branches: one branch for feature extraction, utilizing a depth-wise convolution followed by an atrous window SS2D (AWSS2D) layer, and the other for computing the multiplicative gating signal, which is achieved through a linear mapping followed by an activation layer. Our method integrates an atrous window selective scan mechanism with adjustable rates, facilitating the gradual expansion of the scanning range. This design allows the model to capture both detailed local features and global context, thereby improving its performance in change detection tasks.

As illustrated in Fig. 6 (b), the process of passing data through the AWSS2D module consists of three steps: atrous window scanning (AWScan), selective scanning utilizing S6 blocks, and atrous window scan merging (AWSMerg). The combination of atrous window scanning and atrous window merging is collectively referred to as the atrous window scan module (AWSM). Given a feature map from the encoding phase, AWSS2D first employs four different rates (e.g., 2, 5, 7, 9) to segment the image into four groups of windows with varying sizes. Within each group, image patches from all windows are unfolded and arranged into sequences along a horizontal traversal path (as shown in Fig. 1). These four groups of patch sequences are then processed in parallel using independent S6 modules, with each scanning window independently captures relevant information. This atrous window progressive scanning method ensures a thorough analysis within each selective scan block, striking a balance between global context and local details. Finally, the processed sequences are reshaped and merged using the AWSMerge to generate the output feature map.

To enhance the interaction between channels, we propose an AWSMerge to recalibrate the weights of the feature maps, selectively activating the most relevant change features from the multi-window scan features. As shown in Fig. 6 (c), the spatial dimensions of $X \in R^{h \times w \times c}$ are reduced to a vector $Y \in R^c$ through a global average pooling operation. Therefore, each element in the vector contains global information about its corresponding channel. Specifically, the calculation formula for the $i$-th element in the vector is as follows:

$$Y_i = f_{gap}(X_i) = \frac{1}{H \times W} \sum_{i=1}^{H} \sum_{j=1}^{W} X_i(i,j) \tag{5}$$

Next, the weight vector $V \in R^c$ is generated through two consecutive fully connected layers. The formula for this process is as follows:

$$V = Sigmoid(f_2(ReLU(f_1(Y)))) \tag{6}$$

where $f_1$ and $f_2$ represent two fully connected layers, respectively. Finally, the weight vector V is element-wise multiplied with the input feature Y to produce the final output tensor $\overline{X}$. The operation can be expressed as follows:

$$\overline{X} = V \odot Y \tag{7}$$

The AWSMerge calculates the importance weight of each feature channel, allowing the network to automatically amplify the response to important features while suppressing less relevant ones.

### 3.4. Change detection decoder

#### 3.4.1. The binary change detection decoder

It's significant to learn the spatiotemporal relationships of multi-temporal images for the CD task. As illustrated in Fig. 4, the specific structure of the change decoder is built upon the proposed AWVSS block. By thoroughly exploiting the spatiotemporal correlation information from the feature maps across the four stages of the encoder, high-precision binary change detection results are ultimately produced. We initially rearrange the pre- and post-temporal feature maps obtained from the encoding stage(Chen et al., 2024a). The calculation procedure is described as follows:

$$Z^1 = concat(Z^{T_1}, Z^{T_2}) \tag{8}$$

$$Z^2[b,h,c,w] = \begin{cases} Z^{T_1}[b,h,c,w] & if \ w < W \\ Z^{T_2}[b,h,c,w-W] & if \ w \geq W \end{cases} \tag{9}$$

$$Z^3[b,h,c,w] = \begin{cases} Z^{T_1}[b,h,c,m] & if \ w = 2m \\ Z^{T_2}[b,h,c,m] & if \ w = 2m+1 \end{cases} \tag{10}$$

Subsequently, the AWVSS block is applied to independently scan $Z^1$, $Z^2$ and $Z^3$, with each module effectively capturing both fine-grained local features and global context. In the next step, the results from the three scans are concatenated, followed by channel adjustment and feature extraction through a convolutional layer. Finally, the output features from the current stage are then fused with the feature maps from the previous stage, followed by upsampling before being passed to the next stage.

#### 3.4.2. The semantic change detection decoder:

The architecture of the proposed AWMambaSCD is illustrated in Fig. 5. AWMambaSCD consists of an encoder with two branches and a decoder with three branches. Inspired by the U-Net architecture, the AWVSS block is employed at the beginning of each decoding stage to capture both the local and global spatial context of the input data. Subsequently, the feature maps are upsampled and combined with the information from lower-level feature maps of higher resolution. Finally, the resulting feature maps are smoothed through a residual layer. The calculation steps are as follows:

$$F_4 = AWVSS(Conv Block(X_4)) \tag{11}$$

$$F_3 = ResBlock(AWVSS(UP(ConvBlock(X_3), F_4))) \tag{12}$$

$$F_2 = ResBlock(AWVSS(UP(ConvBlock(X_2), F_3))) \tag{13}$$

$$F_1 = ResBlock(AWVSS(UP(ConvBlock(X_1), F_2))) \tag{14}$$

### 3.5. Loss function

#### 3.5.1. Loss for BCD:

Since BCD can be formulated as a pixel-wise binary classification problem, we adopt a classification loss function to guide the network training. In particular, the binary cross-entropy (BCE) loss is employed to provide pixel-level supervision by measuring the discrepancy between the predicted change map and the ground truth (GT). The BCE loss is defined as follows:

$$L_{cd} = -Y_c \log(P_c) + (1-Y_c)\log(1-P_c) \tag{15}$$

where $P_c$ denotes the predicted probability and $Y_c$ denotes the ground truth label.

#### 3.5.2. Loss for SCD

The semantic segmentation loss is optimized using the multi-class cross-entropy loss function, which measures the discrepancy between the predicted semantic labels and the ground truth. It is formally defined as follows:

$$L_{ss} = -\frac{1}{N}\sum_{i=1}^{N}[Y_i \log(P_i)] \tag{16}$$

where $N$ is the number of semantic classes, and $Y_i$ and $P_i$ denote the ground truth label and the predicted probability for the $i$-th class, respectively.

To enforce semantic consistency between the two predicted semantic segmentation maps, we introduce a semantic change loss formulated under the principles of contrastive learning. It is computed as follows:

$$L_{sc} = \begin{cases} 1 - \cos(X_1, X_2), & Y_c = 0 \\ \cos(X_1, X_2), & Y_c = 1 \end{cases} \quad (17)$$

where $Y_c$ represents the ground truth label, and $X_1$ and $X_2$ refer to the pixel vectors of these two semantic segmentation maps, respectively.

The total loss is calculated as:

$$L_{scd} = \lambda_1 L_{cd} + \lambda_2 L_{sc} + \lambda_3 (L_{ss_1} + L_{ss_2}) \quad (18)$$

## 4. Experiments and analysis

In this section, we evaluate the performance of the proposed model by comparing it with other methods across six datasets.

### 4.1. Dataset Description

**CLCD**: The CLCD dataset (Liu et al., 2022) was acquired by the GF-2 satellite over Guangdong Province, China, in 2017 and 2019, with spatial resolutions ranging from 0.5 m to 2 m. It comprises 600 pairs of cropland change samples, covering diverse land categories such as buildings, roads, lakes, and bare soil. Each sample includes two 512×512 bi-temporal images along with a corresponding binary annotation indicating cropland change. For model training and evaluation, the images were further partitioned into 256×256 patches, yielding 1440 patches for training, 480 for validation, and 480 for testing.

**SYSU-CD**: The SYSU-CD (Shi et al., 2021) consists of 20,000 pairs of aerial images, each with a spatial resolution of 0.5 meters and a fixed size of 256×256 pixels, captured over Hong Kong between 2007 and 2014. This dataset is particularly challenging due to the presence of shadow interference and spatial misalignment, which frequently occur in urban and coastal environments. It focuses on various complex change types, including high-rise buildings, infrastructure development, vessels, roads, and vegetation. The dataset is divided into training, validation, and test subsets with a 6:2:2 split ratio.

**WHUCD**: The WHU-CD dataset (Ji et al., 2018) is a publicly accessible benchmark focused on building change detection. It comprises a pair of high-resolution aerial images, each with a spatial resolution of 0.2 meters per pixel and dimensions of 32,507 × 15,354 pixels, capturing building changes near Christchurch, New Zealand, between April 2012 and 2016. To facilitate model training and evaluation, these large-scale image pairs are preprocessed by cropping into non-overlapping patches of 256 × 256 pixels. The resulting patch pairs are then partitioned into training, validation, and test sets, containing 5,204, 743, and 1,487 samples, respectively.

**SECOND**: The SECOND dataset (Yang et al., 2021) consists of 2,968 pairs of images, each with a resolution ranging from 0.5 to 3 meters and a spatial dimensions of 512×512 pixels. It includes a single "no-change" class along with six land cover categories: water, ground, low vegetation, trees, buildings, and playgrounds. To facilitate processing, the large-scale image pairs are cropped into non-overlapping patches, each with a spatial size of 256×256 pixels. These patch pairs are then divided into training, validation, and test sets, containing 11,872, 3,388, and 3,388 samples, respectively.

**Landsat-SCD**: The Landsat SCD dataset (Yuan et al., 2022) is constructed from Landsat imagery acquired between 1990 and 2020, covering the Tumushuke region in Xinjiang, China, near the edge of the Taklimakan Desert. It comprises 8,468 pairs of images, each with a spatial resolution of 30 meters and dimensions of 416×416 pixels. After excluding samples generated by spatial augmentation methods such as flipping, masking, and resizing, the dataset retains 2,425 pairs of original images. These original samples are divided into training, validation, and test sets containing 1,455, 485, and 485 pairs, respectively.

**JL1**: The JL1 dataset (Wang et al., 2024b) is a competition dataset specifically curated for cropland change detection. The imagery is acquired from the Jilin-1 remote sensing satellite, with each image having a spatial resolution of 0.75 meters per pixel and dimensions of 256×256 pixels in RGB format. This dataset provides nine distinct "from-to" change categories for the bi-temporal image pairs, explicitly labeling the type of land-cover transitions between acquisition dates. The dataset is divided into training, validation, and test subsets, containing 4,050, 1,950, and 1,950 samples, respectively.

### 4.2. Experimental details

All experiments were conducted on an Ubuntu system equipped with an NVIDIA Tesla A40 GPU with 48 GB of memory. The proposed architectures were implemented using PyTorch. For all datasets except Landsat, image pairs and their corresponding labels were cropped to 256 × 256 pixels before being fed into the network. Each dataset was further partitioned into training, validation, and test sets. During training, the network was optimized using the AdamW optimizer with a learning rate of 1e-4 and a weight decay of 5e-3, without employing any learning rate decay strategy. The batch size was set to 8, and the number of training iterations was fixed at 50,000 across all six datasets. To enhance model generalization, data augmentation techniques, including random rotation, left-right flipping, and top-bottom flipping, were applied during training. To promote reproducibility and encourage further research within the community, our source code will be made publicly available.

### 4.3. Evaluation metrics

To comprehensively evaluate model performance across different CD subtasks, we employ a suite of metrics specifically designed to address the distinct requirements of both BCD and SCD tasks. For the BCD subtask, the evaluation metrics include:

$$OA = TP + TN/(TP + FP + TN + FN) \quad (19)$$
$$IoU = TP/(TP + FP + FN) \quad (20)$$
$$Prec = TP/(TP + FP) \quad (21)$$
$$Rec = TP/(TP + FN) \quad (22)$$
$$F1 = 2/(Prec^{-1} + Rec^{-1}) \quad (23)$$

where TP, FP, TN, and FN denote the counts of true positives, false positives, true negatives, and false negatives, respectively. For the SCD task, in addition to the previously mentioned F1 score, evaluation also includes metrics such as mIoU and SeK. Their calculation methods are defined as follows:

$$mIoU = (IoU_0 + IoU_1) \times 0.5 \quad (24)$$

where $IoU_0$ and $IoU_1$ represent the intersection over union (IoU) for unchanged and changed regions, respectively.

$$SeK = \exp(IoU_1 - 1) \cdot (\hat{\rho} - \hat{\eta})/(1 - \hat{\eta}) \quad (25)$$

with

Table 1
Performance comparison of different methods on the CLCD, SYSU-CD, and WHU-CD datasets.

| Type | Method | Params | Flops | CLCD | | | SYSU-CD | | | WHU-CD | | |
|---|---|---|---|---|---|---|---|---|---|---|---|---|
| | | | | OA | IoU | F1 | OA | IoU | F1 | OA | IoU | F1 |
| CNN | ICIF-NET | 23.80 | 25.40 | 95.72 | 50.09 | 66.74 | 89.44 | 62.27 | 76.75 | 98.99 | 78.22 | 87.78 |
| | GAS-Net | 23.50 | 290.0 | 94.73 | 51.10 | 62.28 | 89.24 | 62.86 | 77.18 | 98.93 | 78.37 | 85.15 |
| | CF-GCN | 13.60 | 44.10 | 94.39 | 39.79 | 56.93 | 91.24 | 66.17 | 79.64 | 99.10 | 80.22 | 89.03 |
| | ChangeCLIP* | 120.80 | 41.50 | 96.62 | 61.00 | 75.78 | 92.08 | 70.53 | **83.32** | 99.42 | 86.99 | 93.04 |
| Trans | ChangeFormer | 41.00 | 202.80 | 94.78 | 43.44 | 60.57 | 90.34 | 63.40 | 77.60 | 98.97 | 77.51 | 87.33 |
| | TransUNetCD | 35.96 | 27.68 | 96.50 | 62.40 | 76.85 | 91.51 | 68.75 | 81.48 | 99.21 | 82.67 | 90.51 |
| | SwinSUNet | 42.95 | 12.02 | 96.29 | 58.84 | 74.09 | 90.44 | 63.40 | 77.60 | 99.28 | 84.54 | 91.62 |
| Mamba | MambaBCD_T | 30.35 | 18.67 | 97.08 | 64.82 | 78.66 | 92.09 | 69.77 | 82.20 | 99.42 | 87.14 | 93.13 |
| | AWMambaBCD_T | 30.55 | 18.81 | **97.14** | **65.58** | **79.21** | **92.36** | **70.76** | 82.88 | **99.48** | **88.30** | **93.79** |
| | CD-Lamba* | 28.74 | 15.26 | 96.78 | 64.02 | 78.06 | 91.72 | 70.44 | 82.66 | 99.32 | 86.07 | 92.51 |
| | CD_Mamba | 11.91 | 47.01 | 94.63 | 48.83 | 65.62 | 90.12 | 64.91 | 78.72 | 99.12 | 80.90 | 89.44 |
| | MambaBCD_S | 49.94 | 28.70 | 97.11 | 65.55 | 79.19 | 92.32 | 70.15 | 82.45 | 99.45 | 87.75 | 93.48 |
| | AWMambaBCD_S | 50.15 | 28.85 | **97.33** | **68.35** | **81.20** | **92.51** | **70.94** | 83.00 | **99.49** | **88.32** | **93.80** |

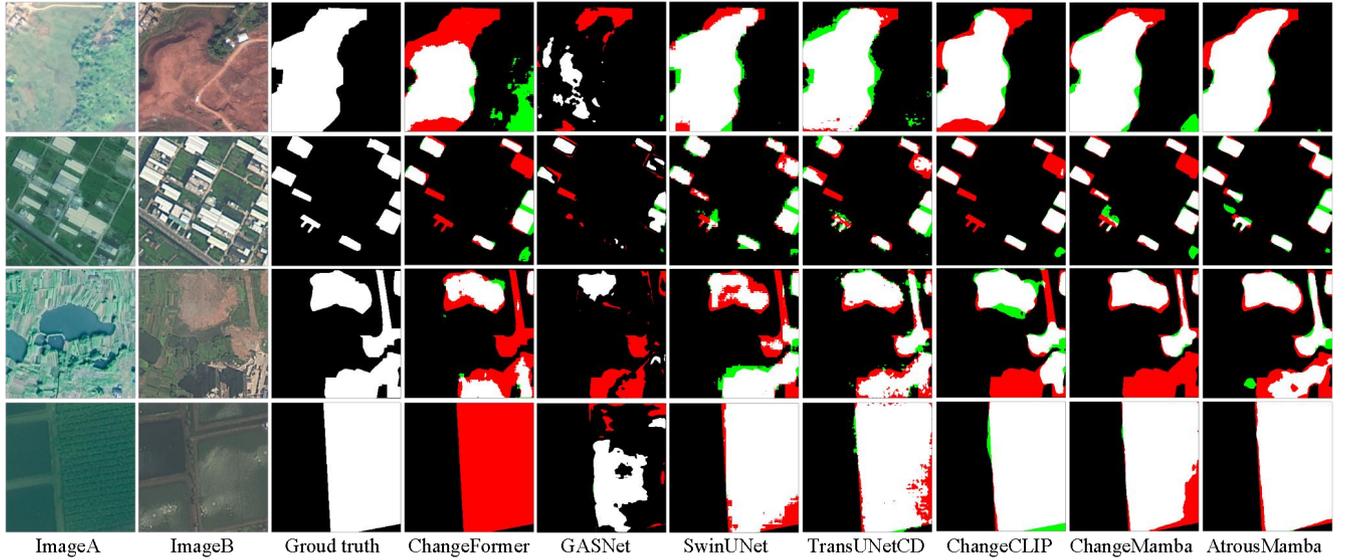

ImageA    ImageB    Groud truth    ChangeFormer    GASNet    SwinUNet    TransUNetCD    ChangeCLIP    ChangeMamba    AtrousMamba

Fig. 7. Visualization results of different models on the CLCD. white, black, green and red represent TP, TN, FP, and FN, respectively.

$$\hat{\rho} = \sum_{i=2}^{C} q_{ii} \Big/ \left( \sum_{i=1}^{C}\sum_{j=1}^{C} q_{ij} - q_{00} \right) \tag{26}$$

$$\hat{\eta} = \sum_{j=1}^{C} \left( \hat{q}_{j+} \cdot \hat{q}_{+j} \right) \Big/ \left( \sum_{i=1}^{C}\sum_{j=1}^{C} q_{ij} - q_{00} \right) \tag{27}$$

where $q_{ij}$ represents the number of pixels classified as the $i$-th change type but actually belonging to the $j$-th change type, and $\hat{q}_{i+}$ and $\hat{q}_{+i}$ denote the row sum and column sum of the confusion matrix, excluding $q_{00}$.

### 4.4. Benchmark methods

To evaluate the superiority of our proposed AtrousMamba, we compared it with several recent state-of-the-art CNN-based, Transformer-based, and Mamba-based algorithms. For BCD, the methods compared include, CF-GCN (Wang et al., 2024c), ICIF-NET (Feng et al., 2022), GAS-Net (Zhang et al., 2023), ChangeFormer (Bandara and Patel, 2022), TransUNetCD (Li et al., 2022), SwinSUNet (Zhang et al., 2022), ChangeCLIP (Dong et al., 2024), MambaBCD (Chen et al., 2024a), CDMamba (Zhang et al., 2024) and CD-Lamba (Wu et al., 2025). For SCD, the methods compared include SSCDl (Ding et al., 2022), BiSRNet (Ding et al., 2022), DEFO-MTLSCD (Li et al., 2024), SCanNet (Ding et al., 2024), CdSC (Wang et al., 2024b), and MambaSCD (Chen et al., 2024a). We trained, validated, and tested these methods on our preprocessed datasets using the hyperparameters, data augmentation strategies, and loss functions specified in their respective original papers.

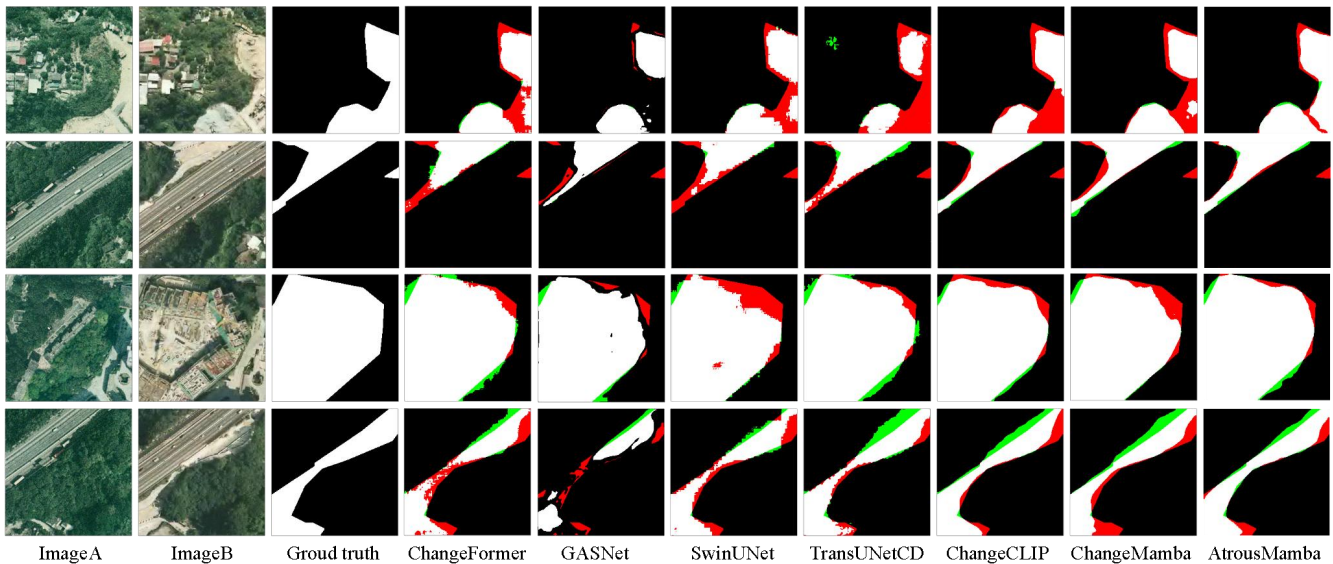

**Fig. 8.** Visualization results of different models on the SYSU-CD. white, black, green and red represent TP, TN, FP, and FN, respectively.

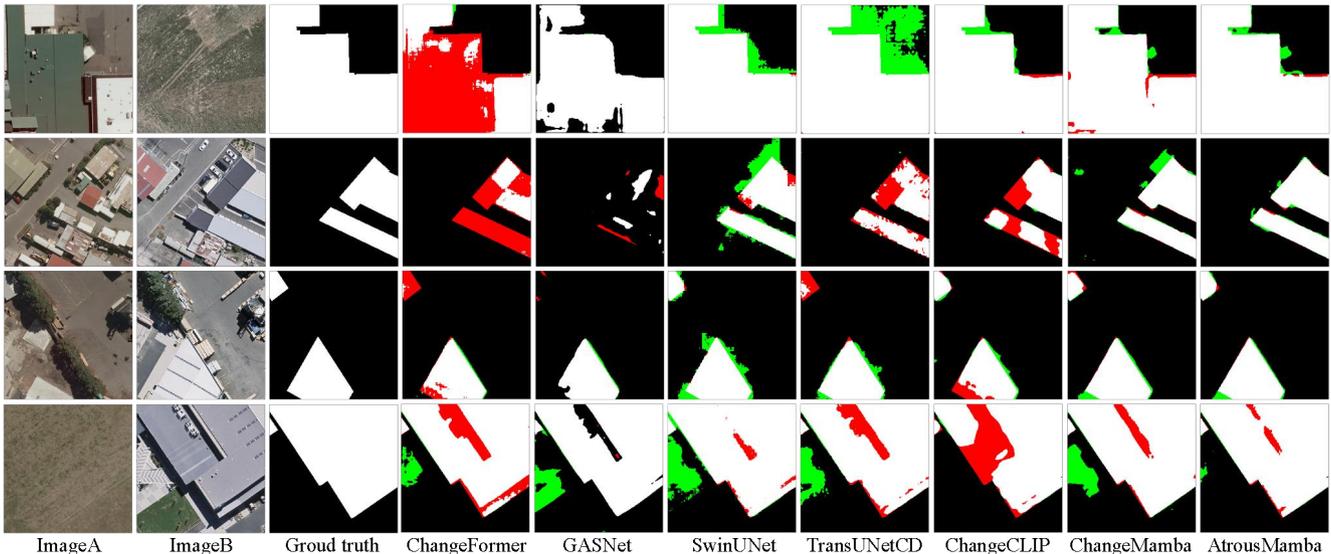

**Fig. 9.** Visualization results of different models on the WHUCD. white, black, green and red represent TP, TN, FP, and FN, respectively.

**Table 2**
Performance comparison of different methods on the SECOND, Landsat-SCD, and JL1-SCD datasets.

| Type | Method | Params | Flops | SECOND | | | Landsat-SCD | | | JL1-SCD | | |
|---|---|---|---|---|---|---|---|---|---|---|---|---|
| | | | | F1 | mIoU | SeK | F1 | mIoU | SeK | F1 | IoU | SeK |
| CNN | SSCDl | 23.31 | 47.44 | 61.45 | 72.55 | 22.11 | 81.68 | 82.60 | 43.52 | 66.25 | 69.85 | 23.55 |
| | BiSRNet | 23.40 | 47.54 | 60.90 | 72.24 | 21.58 | 80.48 | 81.72 | 41.37 | 65.95 | 69.72 | 23.33 |
| | DEFO-MTLSCD | 26.02 | 100.27 | 61.71 | 73.07 | 22.65 | 80.75 | 83.25 | 43.42 | 64.00 | 69.34 | 21.69 |
| Trans | SCanNet | 27.90 | 66.24 | 55.27 | 68.06 | 16.09 | 83.69 | 84.72 | 48.16 | 64.05 | 68.10 | 20.69 |
| | CdSC | 33.85 | 33.70 | 63.69 | 73.46 | 24.43 | 84.01 | 84.81 | 48.57 | 85.05 | 83.48 | 51.86 |
| Mamba | MambaSCD_T | 34.69 | 26.66 | 63.04 | 72.87 | 94.80 | 87.63 | 88.09 | 57.24 | 90.21 | 88.55 | 64.46 |
| | AWMambaSCD_T | 35.03 | 26.86 | **63.86** | **73.62** | **24.70** | **88.05** | **88.32** | **58.07** | **90.13** | **88.75** | **64.67** |
| | MambaSCD_S | 54.28 | 36.70 | 63.40 | 73.15 | 24.04 | 88.38 | 88.71 | 59.10 | 90.39 | 88.92 | 65.26 |
| | AWMambaSCD_S | 54.63 | 36.89 | **64.24** | **73.66** | **24.95** | **89.03** | **89.07** | **60.43** | **90.90** | **89.26** | **66.37** |

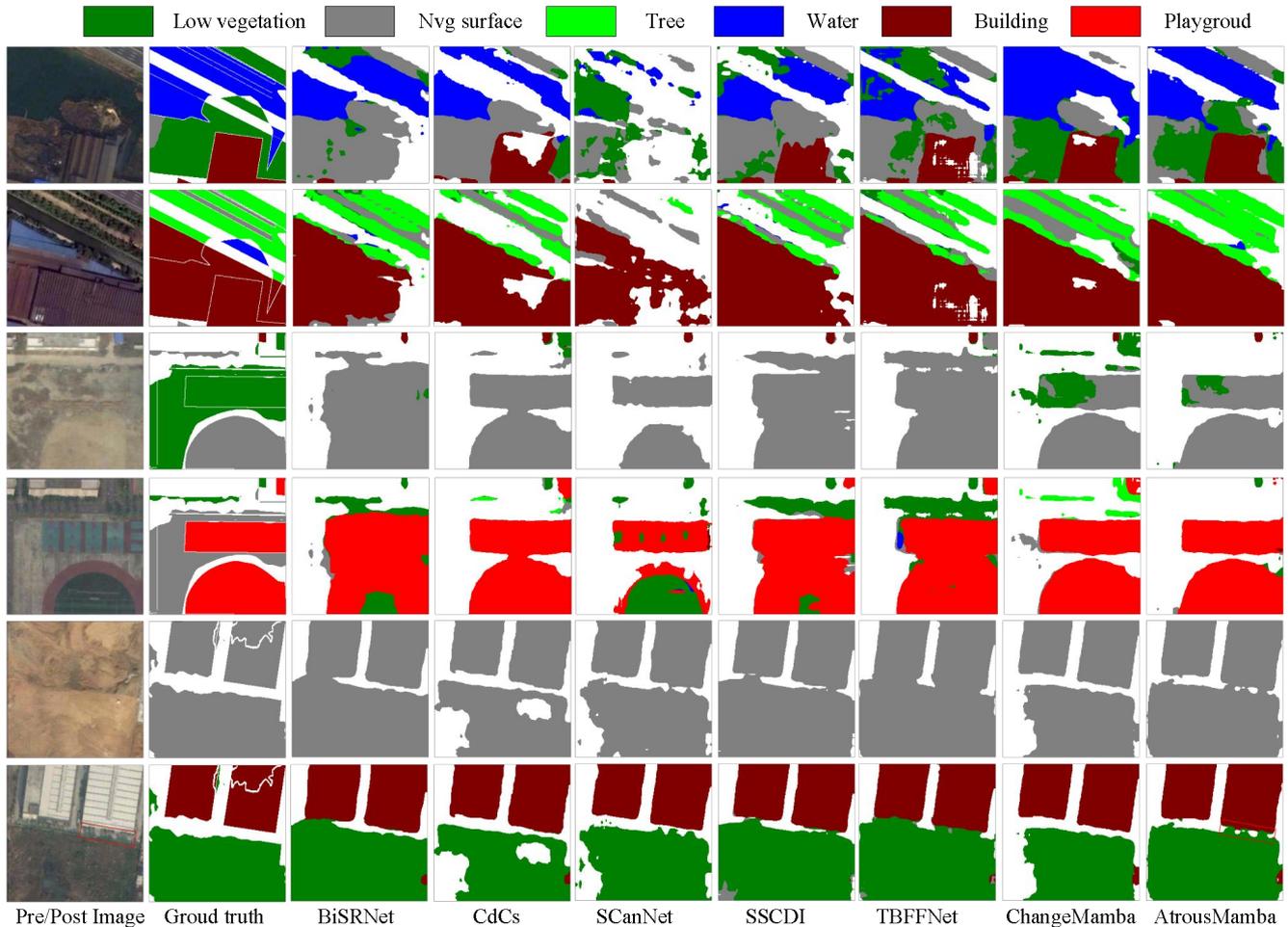

**Fig. 10.** Visualization results of different models on the SECOND.

### 4.5. Benchmark comparison in three BCD subtasks

As shown in Table 1, we compare AWMambaBCD with existing methods on three high-resolution datasets: CLCD, SYSU-CD, and WHU-CD. Here, "CNN" refers to CNN-based models, "Trans" refers to Transformer-based models, and "Mamba" refers to Mamba-based models. All methods compared in this study utilize hierarchical backbone networks, such as CNN, Transformer, and Mamba, as feature extractors. It can be observed that our method significantly outperforms CNN-based, Transformer-based, and other Mamba-based methods. The proposed AWMambaBCD achieves the best overall performance, with its small variant, AWMambaBCD_S, yielding the highest OA (97.33% / 99.49%), IoU (68.35% / 88.32%), and F1 score (81.20% / 93.80%) on the cropland CD (CLCD) and single-object CD (WHU-CD) tasks, respectively. For class-agnostic CD (SYSU dataset), It achieved the highest IoU of 70.94% and the highest OA of 92.51%, while its F1 score (83.00%) was slightly lower than that of ChangeCLIP. Additionally, it outperformed recent SSM-based methods such as ChangeMamba, CD_Mamba and CD-Lambda.

Fig. 7 to Fig. 9 show the binary change maps predicted by our method on the test sets of the three datasets. It is evident that the proposed method effectively detects changes across various types, scales, and sizes, and accurately captures the edge details of the changes present in these image pairs. Our AWMambaBCD method rarely produces false alarms or omission alarms. Due to its ability to enhance spatial locality details and global context, it successfully detects changes that other methods fail to identify. These results fully demonstrate the robust feature extraction and representation capabilities of the AtrousMamba architecture for BCD tasks across diverse and complex scenarios.

Table 1 presents a comparison of the efficiency of these BCD methods on the CLCD, SYSU-CD, and WHU-CD datasets. The computational complexity of the networks is assessed based on the number of parameters (Param.) and floating-point operations per second (FLOPs). The AWMambaBCD architecture demonstrates linear computational complexity while effectively capturing both global and local information. The AWMambaBCD_T variant comprises 30.55M parameters and requires 18.81G FLOPs, which are relatively lower than those of CNN- and Transformer-based models. Although its computational advantage over other Mamba variants is modest, AWMambaBCD_T maintains a lightweight and computationally efficient profile while delivering competitive detection performance. Furthermore, the small variant, AWMambaBCD_S, introduces additional model complexity but achieves improved detection accuracy.

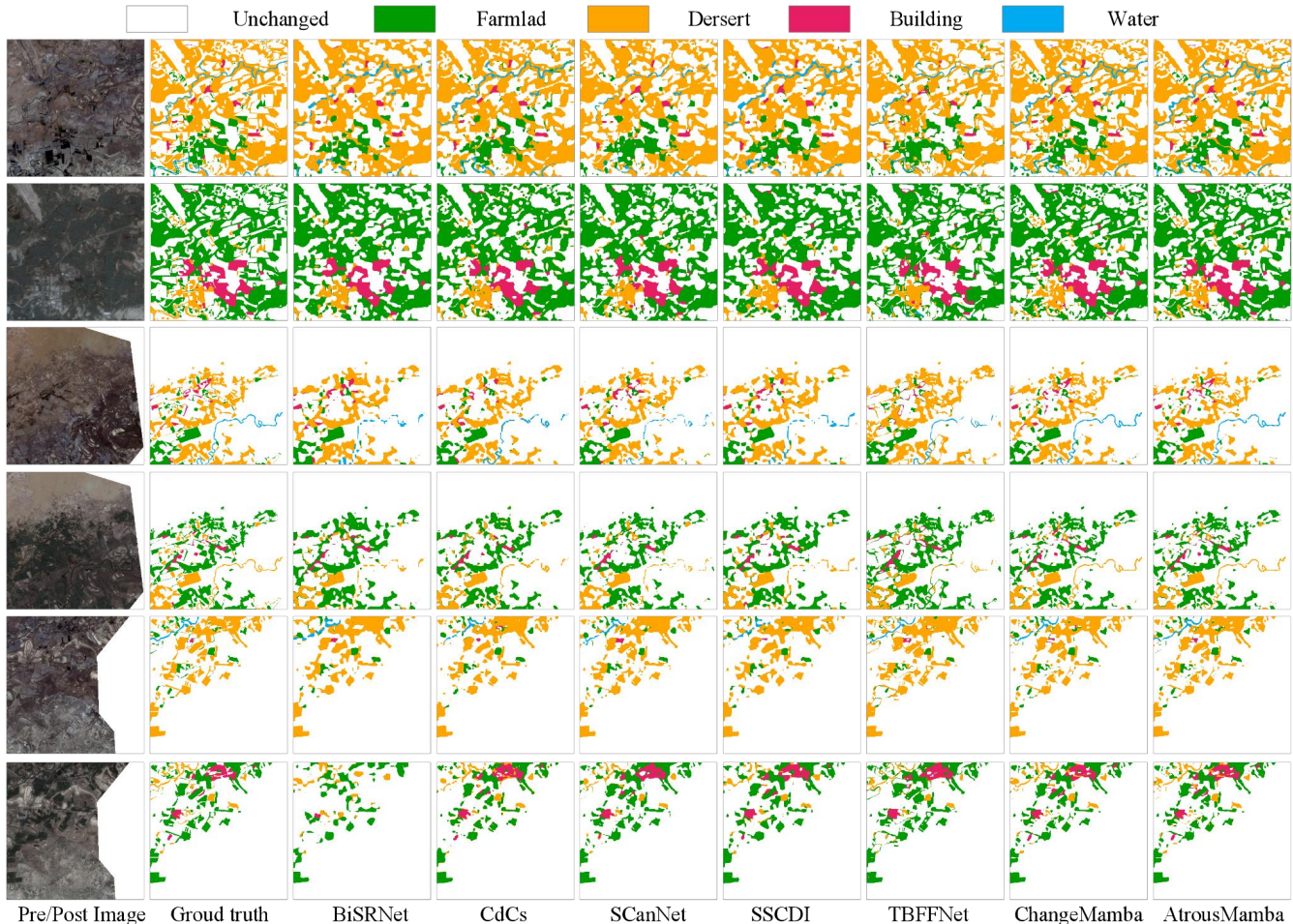

Fig. 11. Visualization results of different models on the Landsat-SCD.

## 4.6. Benchmark comparison in three SCD subtasks

To comprehensively evaluate the performance of the proposed AWMambaSCD, we compared it with seven methods on the SECOND, Landsat, and JL1 datasets. Among these, SSCDl, BiSRNet, and DEFO-MTLSCD are CNN-based methods, while SCanNet and CsSC are Transformer-based, and MambaSCD is based on the Mamba architecture. Except for SSCDl and BiSRNet, all other methods were recently introduced and have demonstrated good performance on remote sensing change detection datasets.

### 4.6.1. SECOND dataset:

The quantitative results are shown in Table 2, The AWMambaSCD_S architecture demonstrates exceptional performance across all four evaluation metrics for SCD tasks. The AWMambaSCD architecture outperforms all SOTA CNN-based, Transformer-based, and Mamba-based methods across all four evaluation metrics for the SCD task. Notably, its small variant, AWMambaSCD_S, achieves the highest performance, with an F1-score of 64.24%, an mIoU of 73.66%, and a SeK of 24.95%.

Fig. 10 shows the visual comparison results of AWMambaSCD and other methods on the SECOND dataset. Our AWMambaSCD effectively identifies and accurately locates various change categories and regions with well-defined boundaries. Especially in scenarios where multiple categories interfere with each other, our AWMambaSCD is better able to distinguish the differences in changes. Even the three small unmarked low-vegetation areas (highlighted in red boxes) in the sixth row are successfully identified. AWMambaSCD effectively reduces the probability of misidentification while generating more complete and accurate representations of land objects.

### 4.6.2. LandSat dataset:

We compared our method with existing approaches on the Landsat-SCD dataset, which consists of medium-resolution images. Our method not only achieves the highest accuracy across all evaluation metrics, but also demonstrates strong adaptability in handling image data with varying resolutions. As shown in Table 2, AWMambaSCD achieves the highest accuracy on the Landsat-SCD dataset, with its small variant, AWMambaSCD_S, yielding 97.37% OA, 89.03% F1 score, 89.07% mIoU, and 60.43% SeK. While CdSC and DEFO-MTLSCD perform well on the SECOND dataset, they lag significantly behind the Mamba-based method, showing lower accuracy. These results highlight the effectiveness of the Mamba architecture on the SCD task and the robustness of our method in handling images varying resolutions.

Fig. 11 visually presents the SCD results of various methods on the Landsat-SCD dataset. It can be observed that our method outperforms others in detecting and localizing semantic change areas related to farmland, water, buildings, and desert. Notably, it excels in accurately identifying fine-scale features, land cover categories, and object boundaries.

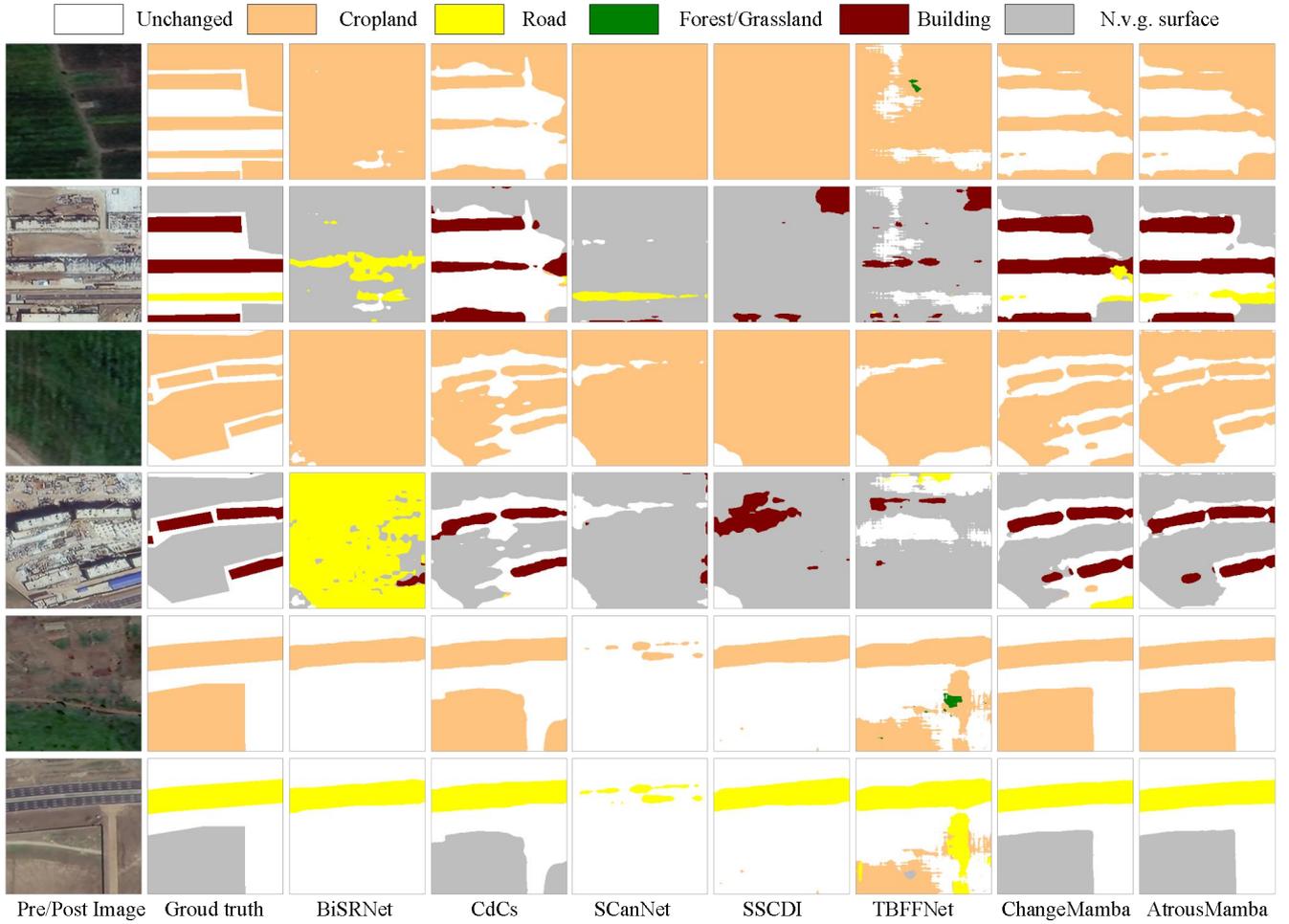

Fig. 12. Visualization results of different models on the JL1-SCD.

**Table 3**
Comparison of different backbone architectures on the CLCD, SYSU-CD, and WHU-CD datasets.

| Method | Params | Flops | CLCD | | | | SYSU-CD | | | | WHU-CD | | | |
|---|---|---|---|---|---|---|---|---|---|---|---|---|---|---|
| | | | OA | IoU | F1 | Prec | OA | IoU | F1 | Prec | OA | IoU | F1 | Prec |
| ResNet50 | 29.30 | 19.95 | 96.70 | 62.48 | 76.91 | 80.30 | 92.20 | 69.17 | 81.77 | 79.86 | 99.12 | 81.64 | 89.89 | 87.88 |
| Swin-Small | 54.60 | 36.72 | 96.95 | 64.20 | 78.19 | 83.40 | 91.19 | 69.27 | 81.85 | 87.61 | 99.39 | 86.53 | 92.78 | 93.96 |
| AWMambaBCD_S | 50.15 | 28.85 | **97.33** | **68.35** | **81.20** | **85.37** | **92.51** | **70.94** | **83.00** | **89.27** | **99.49** | **88.32** | **93.80** | **96.87** |

**Table 4**
Comparison of different backbone architectures on the SECOND, Landsat-SCD, and JL1-SCD datasets.

| Method | Params | Flops | SECOND | | | | Landsat-SCD | | | | JL1-SCD | | | |
|---|---|---|---|---|---|---|---|---|---|---|---|---|---|---|
| | | | OA | F1 | mIoU | SeK | OA | F1 | mIoU | SeK | OA | F1 | mIoU | SeK |
| ResNet50 | 34.40 | 28.31 | 94.61 | 61.97 | 72.57 | 22.80 | 97.09 | 87.17 | 87.62 | 56.00 | 96.34 | 86.15 | 84.72 | 54.62 |
| Swin-Small | 59.35 | 44.72 | **95.58** | 62.47 | 72.73 | 23.16 | 97.04 | 86.45 | 87.10 | 54.42 | 99.15 | 90.41 | 88.77 | 65.03 |
| AWMambaSCD_S | 54.63 | 36.89 | 94.96 | **64.24** | **73.66** | **24.95** | **97.37** | **89.03** | **89.07** | **60.43** | **99.25** | **90.90** | **89.26** | **66.37** |

**Table 5**
Comparison of different scanning strategies on the CLCD, SYSU-CD, and WHU-CD datasets.

| Backbone | Scanning Method | CLCD | | | | SYSU-CD | | | | WHU-CD | | | |
|---|---|---|---|---|---|---|---|---|---|---|---|---|---|
| | | OA | IoU | F1 | Prec | OA | IoU | F1 | Prec | OA | IoU | F1 | Prec |
| VMamba_Small | CSM | 97.10 | 66.81 | 80.10 | 81.80 | 91.88 | 68.94 | 81.62 | 87.49 | 99.47 | 88.02 | 93.63 | 95.32 |
| | Atrous Window Scan | **97.33** | **68.35** | **81.20** | **85.37** | **92.51** | **70.94** | **83.00** | **89.27** | **99.49** | **88.32** | **93.80** | **96.87** |

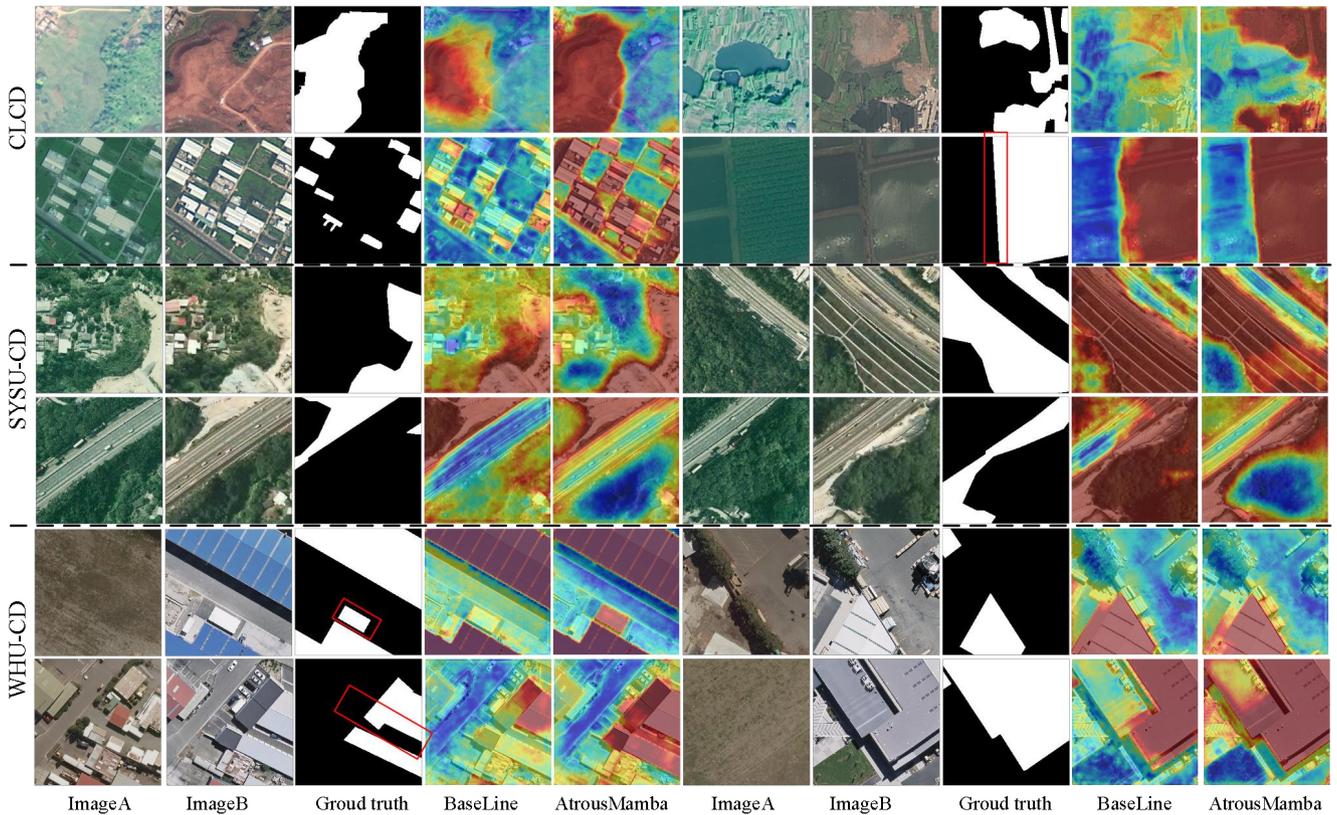

**Fig. 13.** Comparative visualization of heatmaps illustrating different scanning strategies on the CLCD, WHU-CD, and SYSU-CD datasets.

*4.6.3. JL1 dataset*

Table 2 presents the results of various methods on the JL1 dataset with Mamba-based methods significantly outperforming other approaches in terms of quantitative results. AWMambaSCD_S achieved the best performance with an OA of 99.25%, an F1 of 90.90%, an mIoU of 89.26%, and a SeK of 66.37%, demonstrating significant advantages in detecting change regions on the JL1 dataset.

As shown in Fig. 12, the visual comparison results are illustrated to demonstrate the effectiveness of the proposed AWMambaSCD in identifying cropland changes. Although the results of CdSC and ChangeMamba cover the majority of the change regions, they still exhibit some false positives and missed detections. In the second, third, and fourth rows, our method completely detects cropland, buildings, bare soil, and roads, while other methods exhibit incomplete detection or false/missed detections. In the first, fifth, and sixth rows, Mamba-based methods demonstrate a more comprehensive detection of cropland, roads, and bare soil, whereas other methods still suffer from false positives and missed detections.

*4.6.4. Complexity analysis*

Table 2 further lists the number of parameters and FLOPs of these methods. Compared to CNN- and Transformer-based methods, AWMambaSCD_T achieves lower FLOPs of 26.86G, thereby requiring less computational resources while providing faster inference and a more lightweight model. Although AWMambaSCD_T exhibits slightly increased parameter counts, resulting in higher training and storage demands, it significantly surpasses other models in accuracy. Overall, AWMambaSCD_T demonstrates substantial improvements over CNN-, Transformer-, and Mamba-based counterparts, striking a superior balance between performance and efficiency. Moreover, adopting the small variant of AWMambaSCD_S increases model complexity but markedly enhances detection accuracy, especially on the Landsat-SCD and JL1-SCD datasets.

## 5. Ablation

*5.1. Comparison to other backbone networks*

We conducted a comprehensive comparison of Mamba with representative CNN (ResNet) and Transformer (Swin Transformer) backbone networks in both BCD and SCD tasks. As shown in Table 3 and 4, Mamba outperforms both CNN-based and Transformer-based backbone networks on the BCD and SCD, demonstrating its significant advantages as a powerful backbone network for remote sensing change detection. Furthermore, unlike Transformer, whose computational complexity grows quadratically with the input size, Mamba exhibits linear growth in FLOPs while still achieving superior performance, highlighting its advantages in handling larger input scales.

*5.2. Comparison of different scanning methods:*

To further evaluate AtrousMamba's ability to model change-specific features in RSCD tasks, we conducted comparative analyses of different scanning strategies. The baseline model utilizes a global scanning strategy by combining multiple directional passes (Fig. 4), whereas our approach performs localized horizontal scanning within atrous windows (Fig. 1). As illustrated in Table 5, the proposed atrous window scanning strategy decoder further improves the performance of the baseline network, thereby demonstrating AtrousMamba's strong ability to effectively extract fine-grained local features. We provide heatmaps that visually compare different scanning strategies on the CLCD, WHU-CD, and SYSU-CD datasets. As shown in Fig. 13, highlighted regions indicate strong correlations between selected foreground points and other pixels belonging to the same semantic category, reflecting concentrated attention responses. Compared to the baseline, AtrousMamba demonstrates a greater ability to capture foregrounds, changed regions, and global–local contextual interactions, thereby yielding more accurate target localization and sharper boundary delineation. Furthermore, our method is more effective in detecting fine-grained structures and previously unlabeled change areas, further validating the robustness of the proposed atrous window scanning mechanism for RSCD tasks.

## 6. Conclusion

This paper presents AtrousMamba, a novel visual state space model that enhances local dependency modeling while preserving global contextual understanding, specifically tailored for CD tasks. By incorporating an atrous window selective scanning mechanism with adjustable dilation rates, our method enables progressive expansion of the receptive field. Built upon the AWVSS architecture, we propose two dedicated frameworks: AWMambaBCD for BCD and AWMambaSCD for SCD. Additionally, the atrous window state module is introduced to maintain spatial continuity among adjacent tokens and facilitate both global and local receptive field construction, while improving inter-channel feature interactions. Our change decoder is compatible with remote sensing images of varying spatial scales and can be seamlessly integrated into pyramid-based networks. Extensive experiments on six benchmark datasets demonstrate that AtrousMamba achieves SOTA performance compared to CNN-, Transformer-, and Mamba-based approaches. These results highlight its strong generalization ability across multiple CD subtasks and provide new insights into spatio-temporal representation learning for multi-temporal remote sensing imagery.

## CRediT authorship contribution statement

**Tao Wang** Conceptualization, Methodology, Resources, Investigation, Validation, Visualization, Writing - original draft, Data curation, Software, Writing - review & editing. **Tiecheng Bai** Conceptualization, Methodology, Formal analysis, Investigation, Writing - review & editing. **Chao Xu** Investigation, Validation, Formal analysis, Writing - review & editing. **Bin Liu** Investigation, Formal analysis, Writing - review & editing. **Erlei Zhang** Writing - review & editing, Formal analysis, Data curation. **Jiyun Huang** Visualization, Writing - review & editing. **Hongming Zhang** Funding acquisition, Writing - review & editing, Project administration, Resources, Supervision.

## Declaration of competing interest

The authors declare that they have no known competing financial interests or personal relationships that could have appeared to influence the work reported in this paper.

## Acknowledgements

This work was supported by the National Natural Science Foundation of China (42377341) and the Key Research and Development Program of Shaanxi Province(2023-ZDLNY-69).

## Data availability

Data will be made available on request.